\documentclass[runningheads]{llncs}
\usepackage{graphicx}

\usepackage{amsmath,amssymb} 
\usepackage{array,multirow}
\usepackage{ifoddpage}
\usepackage[tableposition=top,labelfont=bf,labelsep=period]{caption}
\usepackage{subcaption}
\usepackage{booktabs, tabularx}
\usepackage{dsfont}
\usepackage{algorithm}
\usepackage{algpseudocode}
\usepackage{algorithmicx}
\usepackage{xspace}
\usepackage[pagebackref,breaklinks,colorlinks]{hyperref}
\usepackage{orcidlink}
\usepackage{colortbl}
\usepackage{bm}
\usepackage[export]{adjustbox}
\definecolor{LightGray}{gray}{0.9}
\definecolor{LightCyan}{rgb}{0.6, 1, 0.996}
\definecolor{LightCyan1}{rgb}{0.6, 1, 0.996}
\definecolor{LightCyan2}{rgb}{0.6, 1, 0.996}
\definecolor{LightMagenta}{rgb}{1, 0.360, 0.768}
\definecolor{LightMagenta1}{rgb}{0.996, 0.682, 0.882}
\definecolor{LightMagenta2}{rgb}{0.858, 0.619, 0.772}
\definecolor{LightOrange}{rgb}{1, 0.772, 0.478}
\definecolor{LightPurple}{rgb}{0.894, 0.741, 0.996}
\definecolor{LightGreen}{rgb}{0.894, 0.996, 0.741}
\definecolor{LightBlue}{rgb}{0.560, 0.623, 1}

\usepackage[capitalize]{cleveref}
\crefname{section}{Sec.}{Secs.}
\Crefname{section}{Section}{Sections}
\Crefname{table}{Table}{Tables}
\crefname{table}{Tab.}{Tabs.}

\makeatletter
\DeclareRobustCommand\onedot{\futurelet\@let@token\@onedot}
\def\@onedot{\ifx\@let@token.\else.\null\fi\xspace}

\def\eg{\emph{e.g}\onedot} 
\def\ie{\emph{i.e}\onedot} 
 
\def\etc{\emph{etc}\onedot} 
\def\wrt{w.r.t\onedot}

\makeatother
\begin{document}
\pagestyle{headings}
\mainmatter
\title{DecisioNet: A Binary-Tree Structured \break Neural Network} 
\titlerunning{DecisioNet}
%
\author{Noam Gottlieb\,\orcidlink{0000-0002-2561-432X} \and
Michael Werman\,\orcidlink{0000-0002-0665-967X}}
\authorrunning{N. Gottlieb \and M. Werman}
%
\institute{The Hebrew University of Jerusalem, Israel\\
\email{\{noam.gottlieb2,michael.werman\}@mail.huji.ac.il}
}
\maketitle              
%

\begin{abstract}
Deep neural networks (DNNs) and decision trees (DTs) are both state-of-the-art classifiers. DNNs perform well due to their representational learning capabilities, while DTs are computationally efficient as they perform inference along one route (root-to-leaf) that is dependent on the input data.
In this paper, we present DecisioNet (DN), a binary-tree structured neural network. We propose a systematic way to convert an existing DNN into a DN  to create a lightweight version of the original model.
DecisioNet takes the best of both worlds - it uses neural modules to perform representational learning and utilizes its tree structure to perform only a portion of the computations.
We evaluate various DN architectures, along with their corresponding baseline models on the FashionMNIST, CIFAR10, and CIFAR100 datasets. We show that the DN variants achieve similar accuracy while significantly reducing the computational cost of the original network.
\end{abstract}

\section{Introduction}
Deep neural networks (DNNs) have achieved exceptional performance in various visual recognition tasks in recent years, such as image classification, object detection, and semantic segmentation. That is mostly due to their representational learning capabilities. However, deploying DNN models in an industrial environment is challenging - especially when the computational resources are low (which is the case for many mobile device applications) or where the model's inference time has to be fast enough (\eg, real-time applications). In addition, a DNN is seen in many cases as a "black box" - one cannot easily figure out \textit{why} a final prediction is made.

Another powerful machine learning model is the Decision Tree (DT). A DT model learns a routing function, where each node of the tree routes the data to one of its children until it reaches a leaf with the final output. This conditional computation means that only part of the DT is visited for each input thus achieving high efficiency. Moreover, DTs provide an interpretable structure, allowing the user to understand why a decision was made. On the downside, DTs usually require hand-engineered data features, and they cannot be trained with gradient-based optimization methods, which limits their expressiveness.

In this paper, we propose a novel general model with the benefits of both DNNs and DTs - the DecisioNet (DN). This is a binary-tree structured DNN, derived from any other DNN that we wish to reduce its computational cost. It can be trained end-to-end using backpropagation \cite{rumelhart1986learning} just like any other DNN. In addition, the DN has routing modules which play the role of the DT within this model, routing the input through the tree. The outcome is a lighter model - in terms of parameters and even more in terms of computational cost - whose performance is at par with the baseline model. We evaluate and compare full baseline models against their DN variants on the FashionMNIST \cite{xiao2017fashion}, CIFAR10, and  CIFAR100 \cite{krizhevsky2009learning} datasets. In this paper, we examine DN only for image classification tasks but this method can be used for other types of tasks as well.

\subsubsection{Contributions} the contributions of this paper are: i) we propose a systematic way of transforming an existing DNN into a tree-structured version of it (its DecisioNet), yielding a lightweight model with comparable performance. ii) we propose a way of training the DN end-to-end, despite the explicit data routing functions within the DN (which involves non-differentiable operations) using  Improved Semantic Hashing. iii) we provide an open-source PyTorch \cite{NEURIPS2019_9015} implementation of DecisioNet, available at \url{https://github.com/noamgot/DecisioNet}.
\section{Related Work}
\subsubsection{Soft Decision Trees} The Soft Decision Tree (SDT) mechanism has been studied in various works, including \cite{suarez1999globally,jordan1994hierarchical,rota2014neural,frosst2017distilling}. SDT is a DT with neural routing nodes sending the data to a sub-tree, multiplied by a probabilistic factor in the range $[0,1]$. The final prediction of the tree is the weighted sum of all leaves, where the weight of each leaf is the   probability of arriving at it. 

This method differs from traditional DTs, whose decisions are binary and deterministic. Another SDT-like method is the Neural Decision Forest (NDF), proposed in \cite{kontschieder2015deep}. This method achieved great performance on the Imagenet dataset \cite{deng2009imagenet}, replacing the last fully-connect layer of a neural network with decision forest nodes. These nodes yield a prediction which is a weighted sum of all the trees' predictions. The major drawback of SDTs is that all paths of the tree must be executed  to get a prediction. These methods lack the advantage of traditional decision trees using only a single computational path, and therefore their efficiency is limited. In this work, we focus on hard decision trees only, which perform only one path of the tree.\\

\subsubsection{Neural Trees With Hard Routing} Another type of neural DT is proposed in \cite{hehn2018end,hazimeh2020tree}, where the forward pass utilizes the routing nodes to make a hard decision, and in this way indeed only the relevant nodes of the tree are visited. \cite{hehn2018end} introduce a routing function which outputs binary values at test time only, such that the tree still performs soft decisions during training, allowing it to train end-to-end. \cite{hazimeh2020tree} on the other hand introduce the Tree Ensemble Layer (TEL), which is capable of performing hard decisions at test time and even during training, by simply skipping unreachable nodes (\ie, nodes whose ancestors reach probability is 0). Notice that the number of reachable leaves is not necessarily limited to 1, hence a soft routing is still performed. In contrast, our method applies fully hard routing (with a single output leaf) both at training time (where we also use soft decision behavior) and at test time. 

\subsubsection{Tree-Structured Neural Networks} Instead of replacing DTs decision nodes with neural modules, the other direction is also possible: creating a DNN with a structure of a DT, such that the routing behavior of a DT is present. This network  consists of multiple sub-networks, that are traversed conditionally based on the routing nodes' decisions. \cite{yan2015hdcnn} proposed HD-CNN, where a small CNN is first activated to classify inputs into coarse categories. This network has a hierarchical architecture which is essentially an SDT-structured DNN with a single decision node with $k$ branches. HD-CNN is not trained end-to-end but in a modular way: there is a common (shared) network for predicting coarse categories and coarse-category expert networks. Each of these networks is trained alone (the expert modules are trained only with their corresponding classes; The coarse-category labels are achieved by clustering similar classes based on an existing network), and all of them are fined tuned together in the end. \cite{ioannou2016decision} extended the tree-structured net of HD-CNN to conditional networks. These are DAG-based CNN architectures with data routing. They distinguish between two types of routing: i) explicit routing - where data is conditionally passed to a node's children (one or more) based on a routing function. This method is similar to DTs. ii) implicit routing - where the data is split into portions that are sent to the node's children unconditionally. In the paper. this method is done using filter groups (\ie, splitting outputs feature maps into groups, where each group goes to a separate route). Another tree-structured NN is Adaptive Neural Trees (ANT) \cite{tanno2019adaptive}. In ANT, the tree structure is learned together with the model's weights; This approach is different from tree-structured NNs with a static architecture that is commonly used (in this work as well).

\paragraph{Differences and Similarities} DecisioNet is, to some extent, a combination of HD-CNN \cite{yan2015hdcnn} and conditional networks \cite{ioannou2016decision}. We create labels for training the routing modules with a method that's based on the one proposed in HD-CNN, but unlike \cite{yan2015hdcnn}, we train deeper, tree-structured DNNs with explicit routing, and we do the training end-to-end. We use explicit routing, similar to the one that was suggested for conditional networks. However, we use a different method for training end-to-end (improved semantic hashing \cite{kaiser2018discrete}), and more importantly, we show actual results for explicitly-routed DNN trees (in \cite{ioannou2016decision} the main results were achieved using implicit routing; The only result that contained explicit routing was a CNN ensemble with a single decision for choosing which models to use for the final prediction). Finally, the approach used by ANT \cite{tanno2019adaptive} is building the tree-structured net from scratch; We aim to optimize the computational cost of an \textit{existing network} by transforming its structure into a tree.


\section{DecisioNet}
\label{sec:decisionet}
The proposed model is based on any existing deep neural network (DNN) such as VGG \cite{simonyan2015deep}, ResNet \cite{he2015deep}, \etc. We'll denote such a DNN as the "baseline model".

\subsection{Architecture} 
\begin{figure}
\centering
\begin{subfigure}[]{.5\textwidth}
    \centering
   \includegraphics[scale=0.3]{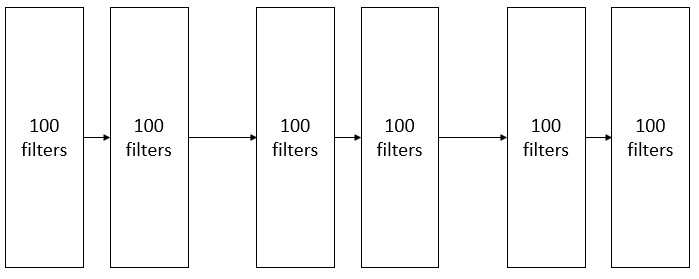}
   \caption{Baseline network}
\end{subfigure}%
\vrule
~
\begin{subfigure}[]{.5\textwidth}
    \centering
   \includegraphics[scale=0.3]{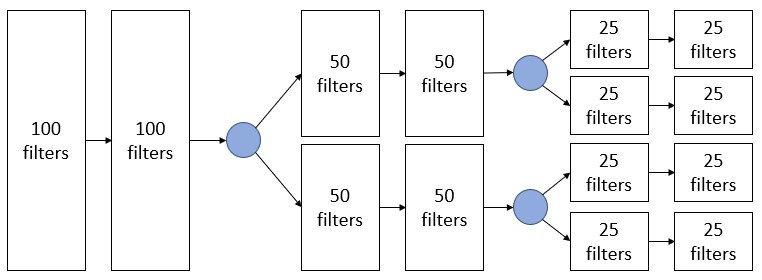}
   \caption{DecisioNet}
\end{subfigure}

\caption{A toy example of transforming a baseline model (a) into its respective DecisioNet (b). For the sake of clarity,  the baseline model consists of 6 convolutional layers, each with 100 filters. The blue circles represent routing modules. In the DN model, the layers after the first split (leftmost blue circle) have half the number of filters compared to their corresponding layers in the baseline model. Similarly, the layers after the second split(s) have a quarter the number of filters. This idea can be easily generalized to an arbitrary number of splits. At test time, each RM picks one of  two routes and the data is passed to the next module through this route only.}
\label{fig:toy_tree}
\end{figure}

The DN architecture is based directly on the baseline model's architecture. It is  a transformation of the baseline model into a binary decision tree, whose primary goal is to achieve similar performance with fewer computational operations. We begin by choosing \textit{split points} - these are the places the new DN model will decide to route the data into one of two routes. In our method, after the $i$-th split (Starting with $i=1$), we split the relevant layers to $2^{i}$ equal portions, in terms of the number of filters in convolutional layers, \etc. A toy example of this idea is displayed in figure \ref{fig:toy_tree}. This way we get a balanced tree whose number of parameters is fewer than that of the baseline model. Moreover, since at test time our model chooses a single path from the root to a leaf, traversing only through a portion of the nodes (just like a DT), the number of multiply-accumulate (MAC) operations is also smaller. The routing at each split is done using a trainable \textit{routing module} (RM). The RM is  a small, lightweight network that makes a binary decision (a detailed explanation is found in section \ref{subsubsec:RM}). 

\subsection{Classes Hierarchical Clustering}
\label{subsec:clustering}
\begin{figure}
\centering
\includegraphics[scale=0.4]{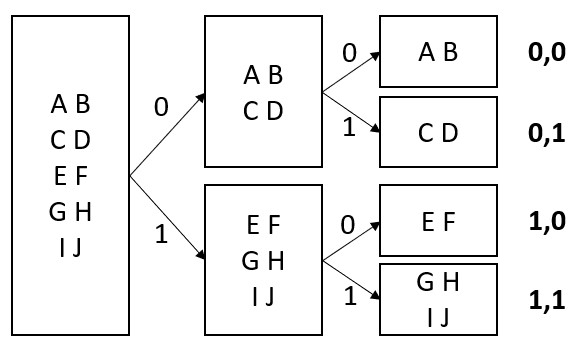}
\caption{Hierarchical clustering example - 10 classes (A-J) are clustered into a clustering tree of depth 2. The deepest (rightmost) clusters share the same routing labels, stated to their right. For example, classes E and F labels are 1 for the first routing, and 0 for the second.}
\label{fig:clustering}
\end{figure}

After deciding the depth of the DecisioNet (\ie, the number of splits), we perform hierarchical clustering of the dataset's classes  to obtain intermediate labels for the routing modules. The goal of this phase is to extract the hierarchical structure of the data, grouping similar classes into nested clusters. The routing labels are derived directly from the  clustering. The general idea is demonstrated in figure \ref{fig:clustering}.

We use a method similar to the one used by \cite{yan2015hdcnn}, extending it to handle more than just 2 hierarchical levels (coarse and fine), using a different clustering method. We begin by randomly sampling a balanced subset of images out of the training set. We evaluate the held-out set using the baseline model to obtain a confusion matrix $\mathbf{F}$. We define a distance matrix $\mathbf{\hat{D}}$ as follows:
\begin{equation}
    \mathbf{\hat{D}}_{ij} = 
    \begin{cases} 
      0 & i = j \\
      1 - \mathbf{F}_{ij} & i \neq j  
   \end{cases}
\end{equation}
To obtain a symmetric distance matrix, we define:
\begin{equation}
    \mathbf{D} = \frac{1}{2}\left(\mathbf{\hat{D}} + \mathbf{\hat{D}}^T\right)
\end{equation}
At this point, $\textbf{D}_{ij}$ measures the similarity between classes $i$ and $j$. Having a distance matrix, we use it to perform hierarchical agglomerative clustering. This phase gives us a division of the dataset's classes into 2 clusters,  a division of each of these clusters into 2 sub-clusters, and so on. This  can be seen as a mapping  of a class to its set of clusters $F_c:\{i\}_{i=1}^{C}\to\{0,1\}^k$, where $C$ is the number of classes and $k$ is the depth of the DecisioNet tree.

Unlike \cite{yan2015hdcnn}, we do not allow overlap between same-level clusters, \ie, at each clustering depth, each class can be found in exactly one sub-cluster. The main problem with this choice is that images that are routed to a wrong branch (at any level, even in the deepest routing module) end up in the wrong leaf and  are miss-classified. We overcome this problem by allowing each leaf of the DN to predict all classes - even classes that shouldn't have been routed to this leaf. Allowing cluster overlap is possible though, and we leave this for  future research.

\subsection{Routing}
One of the key questions when dealing with tree-structured neural networks is which routing method to use. There are 2 main types of routing found in the  literature:
\begin{enumerate}
  \item \textit{Soft routing} - in this method, the output of each branch is multiplied by a real value (usually between 0 and 1) and the final output of some routing point is the sum of its branches' outputs. 
  \item \textit{Hard routing} - in this method, each branch either passes its input or not.
\end{enumerate}

The main advantage of the hard-routing method (and the disadvantage of the soft-routing method), is that it allows us to save computational cost by performing only the chosen path of a given input. To decrease the computational cost, DN uses the hard-routing approach.\footnote{To the best of our knowledge, popular deep learning frameworks, such as PyTorch \cite{NEURIPS2019_9015} and TensorFlow \cite{tensorflow2015-whitepaper}, do not support tree-structured neural nets when using batches of inputs. It means that in practice, during training and evaluating the data passes through all the nodes of the tree where we zero out the "untraversed" ones}

\subsubsection{The Routing Module}
\label{subsubsec:RM}
To choose a route for  input, we use a routing module (RM) - this module is  a small efficient neural network which outputs a single binary value. Inspired by \cite{chen2019look,chen2019selfadaptive}, we used a similar routing module. These papers used this module to choose filters of a convolutional layer, hence they needed multiple outputs; However, we use this module to choose a \textit{single} branch, so we use a slightly modified version of it.

\begin{figure}
\centering
\includegraphics[scale=0.35]{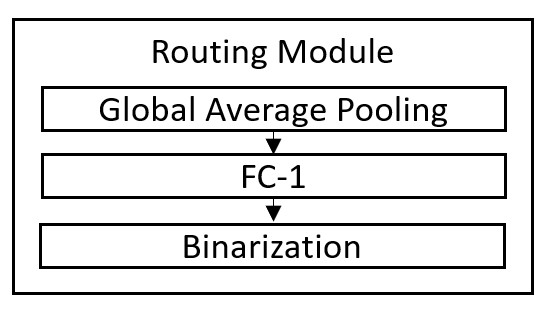}
\caption{A schematic diagram of the proposed routing module. FC-1 is a fully-connected layer with a single output neuron. Full details are in sec. \ref{subsubsec:RM}} \label{fig:rm}
\end{figure}

Let $x \in \mathbb{R}^{C\times H \times W}$ be the output of the last layer before a split in the tree. Our routing module can be defined as follows:
\begin{equation}
\label{eqn:rm}
    RM(x) = B\left(FC\left(GAP\left(x\right)\right)\right)
\end{equation}
Where $GAP$ is a Global Average Pooling operation, defined as:
\begin{equation}
    GAP(x) = \frac{1}{HW} \sum_{i=0}^{H-1}\sum_{j=0}^{W-1}x_{ij}
\end{equation}
$FC$ is a linear projection layer (\ie, a fully-connected layer) with $C$  inputs and a single output and $B$ is a binarization function which will be introduced in the next paragraph. A schematic diagram of the proposed routing module can be found in figure \ref{fig:rm}.

\paragraph{Improved Semantic Hashing}\label{para:ISH} We want our module to output a  binary value - either 0 or 1 -  to choose the next computational path. Performing simple binarization using a threshold is not an option, as this operation is almost always with zero derivative  and therefore we will not be able to perform backpropagation properly. Like \cite{chen2019look,chen2019selfadaptive}, we adopt the Improved Semantic Hashing method, proposed by \cite{kaiser2018discrete}.

During training, we draw a random noise $\epsilon$ with mean 0 and standard deviation 1. We define the following values:
\begin{equation}
    \begin{split}
    & g_\epsilon(z) = z + \epsilon \\
    & g_r(z) =  \sigma ' \left(g_\epsilon\left(z\right)\right) \\ 
    & g_b(z) =  \mathds{1} \left(g_\epsilon\left(z\right) > 0\right)
\end{split}
\end{equation}

$\mathds{1}(z)$ is an indicator function (evaluated to 1 when $z$ is true and 0 otherwise); $\sigma'$ is the saturating sigmoid function from \cite{kaiser2016neural}:
\begin{equation}
    \sigma'(z) = \max\left(0, \min \left(1, 1.2\sigma\left(z\right) - 0.1\right) \right)
\end{equation}
where $\sigma$ is the well-known sigmoid function. Notice that $g_r$ is a real-valued number in the range $[0,1]$ while $g_b$  is a binary value; Moreover, the gradient of $g_r$ is well defined w.r.t $g_\epsilon$, while the gradient of $g_b$ w.r.t $g_\epsilon$ is 0.

Finally, we can define our binarization function $B$ as follows:
\begin{equation}
    B(z) = g_{b/r}(z)
\end{equation}

The term $g_{b/r}$ denotes that in the forward pass we use either $g_r$ or $g_b$, and that is eventually the output of the RM. The choice is done at random with a 50\% chance for each. In the backward pass, we \textit{always} use the gradients of $g_r$  to backpropagate meaningful gradients\footnote{In PyTorch, this operation can be performed like this:\\\texttt{out = g\_b + g\_r - g\_r.detach()}}. To allow this dual behavior, we sum the branches' outputs in the following way:
\begin{equation}
    f(x) = \left(1-\left(RM\left(x\right)\right)\right)f_L\left(x\right) + RM\left(x\right)f_R\left(x\right)
\end{equation}
Where $f_L, f_R$ are the functions applied to $x$ by the left and right branches. Notice that if $RM$ outputs a binary value, only one of the branches is used, while in the real-valued case the branches' outputs are mixed (\ie, soft routing is applied).

During evaluation and inference, we perform the same procedure with 2 changes: first, there is no additive noise ($\epsilon=0$); second, the routing module only outputs binary values, so in this phase, we only use $g_b$ as an output of $RM$.

To conclude, the Improved Semantic Hashing method allows us to train the DecisioNet model in an end-to-end manner and to apply hard decisions at inference time. 

\subsection{Loss Function}
DecisioNet is trained with 2 sets of labels - classification labels and routing labels (extracted in the clustering phase). During training, we want our model to classify inputs correctly while passing classes to their desired routes, based on the routing labels. Therefore, we design a loss function that  balances  these 2 demands. This loss is described as follows:
\begin{equation}
\label{eqn:loss}
    \mathcal{L} = \mathcal{L}_{cls} + \beta \mathcal{L}_\sigma
\end{equation}
Where $\mathcal{L}_{cls}$ is a classification loss (\eg, cross-entropy loss) and $\mathcal{L}_\sigma$ is a MSE loss for the routing labels. $\beta$ is a  hyper-parameter for balancing the classification accuracy and the routing accuracy. We also tried to use a MSE variant which penalizes routing mistakes with different weights depending on the RM depth. The assumption was that it would make sense to give a higher weight to routing mistakes that happen earlier. However, we did not find evidence that this method gives a significant improvement, hence we use the vanilla MSE instead.
\section{Experiments}
\subsection{FashionMNIST \& CIFAR10}
 Fashion-MNIST \cite{xiao2017fashion} consists of a training set of 60K examples and a test set of 10K examples. Each example is a 28$\times$28 grayscale image, associated with a label from 10 classes. The CIFAR10  dataset \cite{krizhevsky2009learning} consists of 60K 32$\times$32 color images in 10 classes, with 6000 images per class. There are 5000 training images and 1000 test images per class. Both datasets were trained using their training images and evaluated on the test images.

\subsubsection{Model}\label{subsubsec:model} 
\begin{figure}[]
\centering
\begin{subfigure}[]{\textwidth}
    \centering
   \includegraphics[width=.85\linewidth]{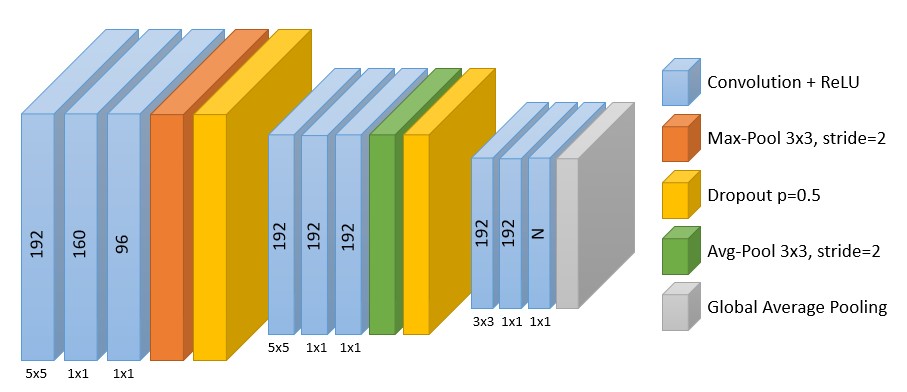}
   \caption{}
   \label{fig:nin_arch}
\end{subfigure}

\begin{subfigure}[]{\textwidth}
    \centering
   \includegraphics[width=.85\linewidth]{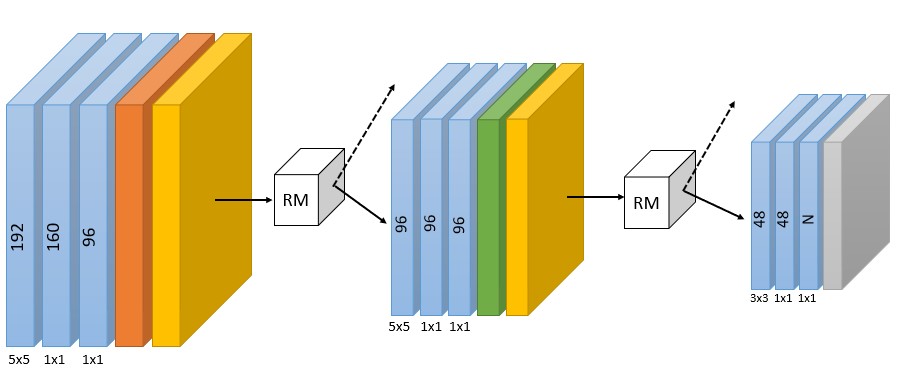}
   \caption{}
   \label{fig:dn_arch}
\end{subfigure}

\begin{subfigure}[]{\textwidth}
    \centering
   \includegraphics[width=.85\linewidth]{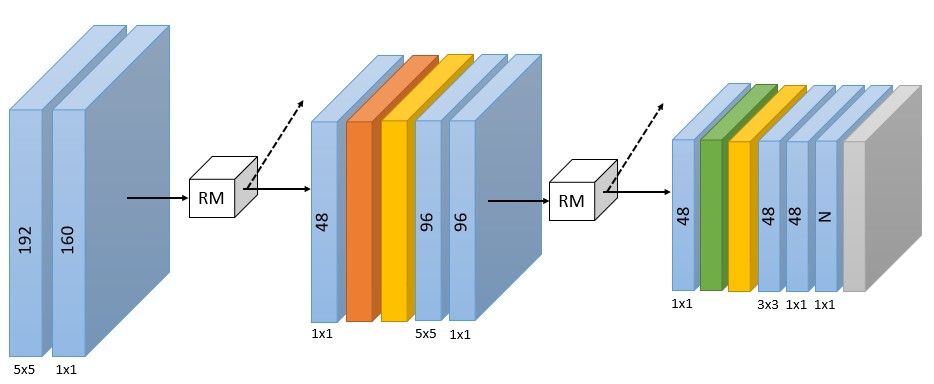}
   \caption{}
   \label{fig:dn_slim_arch}
\end{subfigure}
\caption{In (a) we display the Network-in-Network (NiN) baseline model architecture. For convolutional layer blocks, the kernel size is stated below each block, and the number of filters is written on the side. We always add zero-padding such that the output and input spatial dimensions are the same (only pooling layers reduce the spatial dimensions). $N$ is the number of output classes (in our case $N=10$). In (b) and (c) we display a computational path of 2 DecisioNet variants of the NiN model - DN2 and DN2-slim, respectively. The routing module (RM) is the one that is described in sec. \ref{subsubsec:RM}. Notice that the first block after the RM has  convolutional layers with half the amount of filters compared to the baseline model; After the second RM, the corresponding block contains a quarter of the filters in the convolutional layers.}
\label{fig:arch}
\end{figure}

We used Network-In-Network (NiN) \cite{lin2013network} as our baseline model. The architecture we used is displayed in figure \ref{fig:nin_arch}. We treat it as a stack of 3 small networks (which we refer to as "blocks"). The DecisioNet variants of this model are generated using the method described in section \ref{sec:decisionet}. We experimented with 4 varieties of DecisioNets:
\begin{enumerate}
    \item \textit{DN2} - A basic version with 2 splits, located in-between the baseline NiN model blocks. An example of a single DN2 computational path is displayed in figure \ref{fig:dn_arch}. Since there are 2 splits in the DN2 tree, there are 4 different computational paths (from root to leaf).
    \item \textit{DN1-early} - this version contains a single split point, between the first and second NiN blocks. It is equivalent to the DN2 without the deeper routing modules.
    \item \textit{DN1-late} - this version contains a single split point, between the second and third NiN blocks.  It is equivalent to the DN2 without the first routing module and replacing the 2 deeper RMs with a single one.
    \item \textit{DN2-slim} - A slimmer version of DN2, where we "push" the routing modules to earlier stages compared to DN2. It results in an even lighter version of DN2 (fewer parameters and MAC operations). An example of a single DN2-slim computational path is displayed in figure \ref{fig:dn_slim_arch}.
\end{enumerate}
In all cases, the last layer of the DN model outputs 10 values. It means that classes that reached  the wrong leaf can still be predicted correctly. This way we overcome the aforementioned choice of not allowing overlapping between clusters.

\subsubsection{Preprocessing} For each dataset we calculate its training set mean and standard deviation values (per channel) and use them to normalize the images before feeding them into the model. We also created the routing labels using the method described in \ref{subsec:clustering}. For CIFAR10, we conduct 2 sets of experiments - one with data augmentation (which includes random flips and crops, zero-padded by 4 pixels from each side) and one without it. The data-augmented experiments are marked in the tables with a "+" sign.

\subsubsection{Training} We apply a similar training method to the one used in \cite{lin2013network}: we used stochastic gradient descent with mini-batches of 128 samples, momentum of 0.9, and weight decay of 0.0005. The initial learning rates of FashionMNIST and CIFAR10 were 0.01 and 0.05, respectively. We decrease the learning rate by a factor of 10 when there are 10 epochs without training accuracy improvement. We repeat this process twice during the training period. The training stops if either we reach 300 epochs, or the test loss stops improving for 30 epochs (the latter is to prevent overfitting).

The DecisioNet  training process is identical and has the additional hyperparameter $\beta$ for the loss function (eq. \ref{eqn:loss}). The value of $\beta$ varies between datasets and models. The specific value that was used for every experiment appears next to the results (table \ref{tab:results}).

We initialize the weights of the convolutional layers from a zero-centered Gaussian distribution with a standard deviation of 0.05.

\subsubsection{Clustering Interpretability} Aside from the performance of the different models, we show that the clustering method partitions  the data in an intuitive way. For example, the CIFAR10 first partition is into 2 clusters containing animals and vehicles. The vehicle cluster is then separated into land vehicles (car, truck) and non-land vehicles (plane, ship). In FashionMNIST the first division is into footwear (sandal, snicker, ankle-boot) and non-footwear. The latter is then partitioned into legwear (a singleton cluster with only pants) and non-legwear (T-shirt, dress, bag, etc.). The full hierarchical clustering (for 2 decision levels) is displayed in table \ref{tab:clustering}.

\begin{table}
\caption{Hierarchical clustering results (better viewed in color).}
\label{tab:clustering}
\begin{subtable}{.5\linewidth}\centering
{\begin{tabular}{|c|c|c|} 
        \hline
        Class &  1st level &  2nd level \\
        \hline
        plane &\cellcolor{LightCyan}0 &  \cellcolor{LightOrange}0 \\ 
        ship  &\cellcolor{LightCyan}0 &   \cellcolor{LightOrange}0 \\ \cline{3-3}
        car   &\cellcolor{LightCyan}0 &   \cellcolor{LightPurple}1 \\
        truck &\cellcolor{LightCyan}0 &     \cellcolor{LightPurple}1 \\ 
        \cline{2-3}
        bird  & \cellcolor{LightMagenta}1 &  \cellcolor{LightGreen}0 \\
        cat   & \cellcolor{LightMagenta}1 &  \cellcolor{LightGreen}0 \\
        deer  & \cellcolor{LightMagenta}1 &  \cellcolor{LightGreen}0 \\
        dog   & \cellcolor{LightMagenta}1 &  \cellcolor{LightGreen}0 \\
        frog  & \cellcolor{LightMagenta}1 &  \cellcolor{LightGreen}0 \\\cline{3-3}
        horse & \cellcolor{LightMagenta}1 &  \cellcolor{LightBlue}1 \\
        \hline
  \end{tabular}}
\caption{CIFAR10 clustering}\label{tab:clustering_a}
\end{subtable}%
\begin{subtable}{.5\linewidth}\centering
{\begin{tabular}{|c|c|c|} 
        \hline
        Class &  1st level &  2nd level \\
        \hline
        T-shirt &\cellcolor{LightCyan}0 &            \cellcolor{LightOrange}0 \\
        Pullover &\cellcolor{LightCyan}0&  \cellcolor{LightOrange}0 \\
        Dress &\cellcolor{LightCyan}0 &      \cellcolor{LightOrange}0 \\
        Coat  &\cellcolor{LightCyan}0 &      \cellcolor{LightOrange}0 \\
        Shirt &\cellcolor{LightCyan}0 &     \cellcolor{LightOrange}0 \\
        Bag   &\cellcolor{LightCyan}0 &      \cellcolor{LightOrange}0 \\
        \cline{3-3}
        Trouser     &\cellcolor{LightCyan}0 &       \cellcolor{LightPurple}1 \\
        \cline{2-3}
        Sneaker     &\cellcolor{LightMagenta}1 &  \cellcolor{LightGreen}0 \\
        Ankle-boot  &\cellcolor{LightMagenta}1 &  \cellcolor{LightGreen}0 \\ \cline{3-3}
        Sandal      &\cellcolor{LightMagenta}1 &  \cellcolor{LightBlue}1 \\
        \hline
\end{tabular} }
\caption{FashionMNIST clustering}\label{tab:clustering_b}
\end{subtable}
\end{table}

\subsubsection{Results and Comparison}\label{subsec:rescomp} The results, containing the test accuracy, number of parameters, and number of MACs are presented in table \ref{tab:results}. For each experiment (row in the table) we measured the average accuracy based on 10 runs with an identical setup (except for the initial weights, of course). For the DN variants, the $\beta$ values of eq. \ref{eqn:loss} that were used also appear in the table. The parameters and MAC count were calculated \textit{for a single input image forward pass} using torchinfo \cite{Yep_torchinfo_2020}. A "+" stands for data augmentation. The DN architectures are described in sec. \ref{subsubsec:model}.

\begin{table}[t]
    \centering
    \caption{Accuracy results, along with parameters and MACs count. For experiments with augmented data (marked with "+") we ommit the parameters and MACs count, as they are identical to those of the non-augmented experiments. More details in sec. \ref{subsec:rescomp}.}
    \label{tab:results}
    \LARGE
    \makebox[\textwidth]{\resizebox{1.1\linewidth}{!}{\def\arraystretch{1.5}%
    \begin{tabular}{clccccccc}
    \toprule
    Dataset & Architecture        & $\beta$ &  Acc. (\%) & Acc. Change (\%)    & Params (K) & Params change (\%) & MACs (M) & MACs change (\%)\\
    \midrule
    \multirow{5}{*}{\rotatebox[origin=c]{90}{FashionMNIST}} & Baseline   & - & 92.7 & - &  957.386 & - & 163.3    &  \\
    & DN1-early &  3.0 &  92.9  & $\uparrow$ 0.2  &  736.309 & $\downarrow$23.1 & 93.64  & $\downarrow$42.7\\
    & DN1-late   &  3.0 &  93.4 & $\uparrow$ 0.7  &  939.157 & $\downarrow$1.9 & 153.76 & $\downarrow$5.8 \\
    & DN2      &  3.0 &  92.8 & $\uparrow$ 0.1 &  727.307& $\downarrow$24.0 & 91.24  & $\downarrow$44.1 \\
     & DN2-slim       &   3.0 &  92.7 & 0.0 &  414.027& $\downarrow$56.8 &  60.68  & $\downarrow$62.8 \\
    \midrule
    \multirow{5}{*}{\rotatebox[origin=c]{90}{CIFAR10}} & Baseline        & - &  87.0 & - &  966.986 & - & 223.12   & -\\ 
    &DN1-early  &  3.0 &  86.6         & $\downarrow$ 0.4 &  745.909& $\downarrow$22.9 & 132.14 & $\downarrow$40.8\\
    &DN1-late   &  3.0 &  88.2        & $\uparrow$1.2 & 948.757& $\downarrow$1.9 & 210.66 & $\downarrow$5.6\\
    &DN2        &  0.5 &  86.8       & $\downarrow$ 0.2 & 736.907& $\downarrow$23.8 & 129.00 & $\downarrow$42.2\\
    & DN2-slim        &   1.0 &  86.2 & $\downarrow$ 0.8 &  423.627 & $\downarrow$56.2 & 89.08  & $\downarrow$60.1 \\
    \midrule
        \multirow{5}{*}{\rotatebox[origin=c]{90}{CIFAR10+}}  & Baseline       & - &  88.4 &     -  & - & -   &   -      & -\\
    &DN1-early  &  3.0 &  87.8         &$\downarrow$ 0.6 &  - & - & - & -\\
    &DN1-late   &  3.0 &  89.4        & $\uparrow$ 1.0 & - & - & - & -\\
    &DN2        &  0.5 &  87.6       & $\downarrow$ 0.8 & - & - & - & -\\
    &DN2-slim   &   1.0 &  86.3 & $\downarrow$ 2.1 &  -  & - & - & - \\
    \bottomrule
    \end{tabular}}}
\end{table}

\paragraph{Analysis:} The results (table \ref{tab:results}) show that our method works well for these datasets. The performance of the different DN variants is at par with the baseline NiN model (and sometimes even slightly  outperforms it). Depending on the exact architecture, we  save up to roughly 60\% of the computational cost and memory with a negligible decrease in the model's performance.

Comparing DN models whose routing module is placed earlier in the net (DN1-early and DN2-slim) with their similar alternatives (DN1-late and DN2, respectively), we can see a cost-performance tradeoff: when we put the routing module earlier, the model's performance decreases, along with its cost. This phenomenon is another testimony to the well common belief that shallow layers extract features that are important for all classes.

We also would like to emphasize the significance of choosing an appropriate $\beta$: when setting $\beta=0$, we essentially encourage the DN to only optimize the classification loss, ignoring the routing loss. We found that the performance in this case, is similar to a random choice (\ie, DN outputs a constant prediction). This is surprising because in some cases the DN ignored the routing labels at some point, while still improving its classification accuracy.

\subsection{CIFAR100}
 The CIFAR100 dataset \cite{krizhevsky2009learning} is  like the CIFAR10, except that it has 100 classes containing 600 images each (500 training images and 100 test images per class).
 
\begin{figure}
    \centering
    \begin{minipage}{.45\linewidth}
            \begin{subfigure}[t]{.9\linewidth}
                \centering
                \includegraphics[scale=0.4]{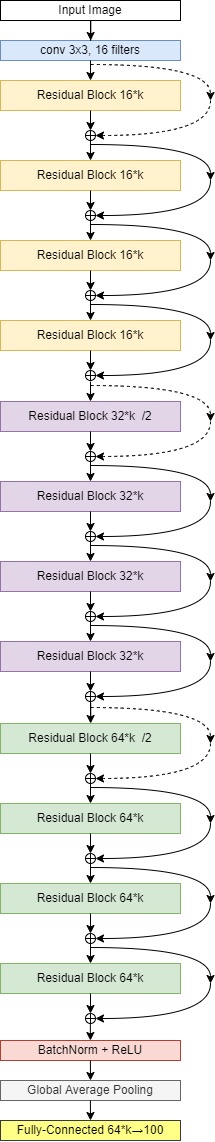}
                \caption{Baseline WRN-n-k architecture with $n=28$. The dashed arrows are 1x1 convolutions for dimensions matching}
                \label{fig:wresnet}
            \end{subfigure}
        \end{minipage}
    \begin{minipage}{.45\linewidth}
        \begin{subfigure}[t]{.9\linewidth}
            \centering
            \includegraphics[scale=0.4]{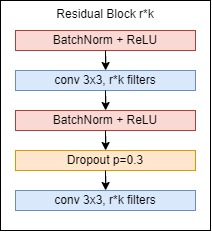}
            \caption{Residual block architecture. In some cases, the first convolution has a stride of size 2 (In (a) these blocks has the additional "/2" text)}
            \label{fig:resblock}
        \end{subfigure}\\[.5cm]
        \begin{subfigure}[b]{.9\linewidth}
        \centering
                \includegraphics[scale=0.45]{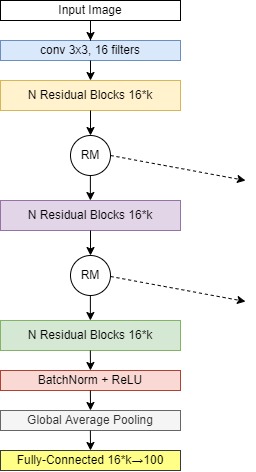}
            \caption{WRN-DN2 single computational branch (we group the residual blocks from (a) into a single block for clarity; Generally, $N=(n-4)/6$, so in our case $N=4$). RM is the routing module.}
            \label{fig:wresnet_dn}
        \end{subfigure} 
    \end{minipage}
    \caption{\textbf{Wide-ResNet-n-k architectures} - in this paper we only use $n=28$ and $k=10$.}
    \label{fig:wresnet_diagrams}
\end{figure}

 For this dataset, we experimented with Wide ResNet (WRN) \cite{zagoruyko2016wide} as our baseline network. We used the version with 28 layers and a depth factor of $k=10$ (denoted as WRN-28-10 in the original paper). We created 3 DN variants out of this network: DN2, whose splits are in between the residual blocks (2 splits), along with DN1-early and DN1-late - each of them with a single split point, positioned at the first or last DN2 split point position. A diagram of the baseline model and the DN2 variant are displayed in figure \ref{fig:wresnet_diagrams}, and the results of our experiments are in table \ref{tab:results_cifar100}.
 
 \subsubsection{Training} We train the network using the same method as in \cite{zagoruyko2016wide}. That is, we use SGD with Nesterov momentum of 0.9 and a weight decay of 0.0005. The
initial learning rate is set to 0.1 and is decreased by a factor of 5 when we reach 60, 120 and 160 epochs. The training is stopped after 200 epochs. The batch size used for training is 128. We  augmented the dataset with the same method as we did with CIFAR10, except padding crops with reflections instead of zeros (that is done to follow the training procedure from the original paper).

 \begin{table}
    \begin{center}
    \caption{CIFAR100 results}
    \label{tab:results_cifar100}
    \makebox[\textwidth]{\resizebox{1.1\linewidth}{!}{\def\arraystretch{1.5}%
    \begin{tabular}{lccccc}
    \toprule
    Architecture        & $\beta$ &  Top-1 Acc. (\%) & Top-5 Acc. (\%)    & Params (M)  & MACs (G) \\
    \midrule
         WRN Baseline       & - &  80.7 &  95.4 & 36.537 &  5.24  \\
    WRN-DN1-early        &  3.0 &  77.6 ($\downarrow3.1$) &        93.9($\downarrow1.5$) & 19.385($\downarrow46.9\%$)   & 2.6 ($\downarrow50.4\%$)\\
    WRN-DN1-late        &  3.0 &   78.7 ($\downarrow2.0$) &        94.5($\downarrow0.9$) & 23.635 ($\downarrow35.3\%$)   & 3.94 ($\downarrow24.8\%$) \\
    WRN-DN2        &  5.0 &  75.7 ($\downarrow5.0$) &        92.7($\downarrow2.7$) & 12.935 ($\downarrow64.6\%$)  & 2.28 ($\downarrow56.5\%$) \\
    \bottomrule
    \end{tabular}}}
    \end{center}
    \bigskip
\end{table}

\paragraph{Analysis:} we see  similar behavior to the  former experiments. The top1 accuracy drop is larger than in the former experiments. One possible explanation to this is the fact that the parameter ratio between the DNs and the baseline model is significantly higher. Different routing module positioning might yield better performance (at some computational cost).

\section{Conclusion}
We introduced DecisioNet (DN), a binary-tree structured network with conditional routing. We proposed a systematic way of building it based on an existing baseline network. DN utilizes Improved Semantic Hashing for training end-to-end, applying conditional routing both during training and evaluation. DN takes advantage of its tree structure, passing inputs only through a portion of the net's neural modules saving a lot of the computational cost. We evaluated multiple DN variants along with their baseline models on multiple image classification datasets and showed that DN is capable of achieving similar performance (\wrt its baseline model) while significantly decreasing the computational cost.

\subsubsection{Future Research}
One clear drawback of our method is the use of additional labels. In addition, these labels are generated using a pre-trained model. Learning hard routing without such auxiliary labels is challenging (as we saw when we set $\beta=0$), and it might be an interesting direction for future research. Another direction is creating DN with unbalanced trees: in this paper, we only examined balanced DNs, while our clustering trees were not necessarily balanced. A possible idea is to divide the computational power of branches proportionally to the size of the data they should process (\eg, a dataset with 100 classes which is split into clusters of size 70 and 30, would have one branch with 70\% of the computational power and the other would have 30\%). 

\subsubsection{Acknowledgements} Thanks to the ISF (1439/22) and the DFG for funding.
\newpage
\bibliographystyle{splncs04}
\bibliography{egbib}

\end{document}